\documentclass[letterpaper]{article} 
\usepackage{aaai2026}  
\usepackage{times}  
\usepackage{helvet}  
\usepackage{courier}  
\usepackage[hyphens]{url}  
\usepackage{graphicx} 
\usepackage{nameref}
\usepackage{natbib}  
\usepackage{caption} 
\usepackage{placeins}

\usepackage{pifont}
\newcommand{\cmark}{\ding{51}}  
\newcommand{\xmark}{\ding{55}}  
\usepackage{bbding}

\usepackage{enumitem}

\usepackage{amsmath}
\usepackage{amssymb}

\usepackage{booktabs}
\usepackage[table]{xcolor}
\definecolor{myblue}{RGB}{230,230,255}
\definecolor{lightgray}{gray}{0.91}
\usepackage{latexsym}
\newcommand*{\belowrulesepcolor}[1]{%
  \noalign{%
    \kern-\belowrulesep 
    \begingroup 
      \color{#1}%
      \hrule height\belowrulesep 
    \endgroup 
  }%
} 
\newcommand*{\aboverulesepcolor}[1]{%
  \noalign{%
    \begingroup 
      \color{#1}%
      \hrule height\aboverulesep 
    \endgroup 
    \kern-\aboverulesep 
  }%
}

\usepackage{subcaption}

\usepackage{algorithm}
\usepackage{algorithmic}

\usepackage{newfloat}
\usepackage{listings}
\DeclareCaptionStyle{ruled}{labelfont=normalfont,labelsep=colon,strut=off} 
\floatstyle{ruled}
\newfloat{listing}{tb}{lst}{}
\floatname{listing}{Listing}
\title{ViCToR: Improving Visual Comprehension via Token Reconstruction for Pretraining LMMs}
\author{
    Yin Xie\textsuperscript{\rm 1}\equalcontrib,
    Kaicheng Yang\textsuperscript{\rm 1}\equalcontrib,
    Peirou Liang\textsuperscript{\rm 2}\equalcontrib,
    Xiang An\textsuperscript{\rm 1},
    Yongle Zhao\textsuperscript{\rm 1}, \\
    Yumeng Wang\textsuperscript{\rm 1},
    Ziyong Feng\textsuperscript{\rm 1},
    Roy Miles\textsuperscript{\rm 3},
    Ismail Elezi\textsuperscript{\rm 3},
    Jiankang Deng\textsuperscript{\rm 4}\thanks{Corresponding Author}
}
\affiliations{
    \textsuperscript{\rm 1}DeepGlint
    \textsuperscript{\rm 2}University of Science and Technology of China \\
    \textsuperscript{\rm 3}Huawei London Research Center 
    \textsuperscript{\rm 4}Imperial College London \\
    
    \{yinxie,kaichengyang,xiangan\}@deepglint.com,jiankangdeng@gmail.com
    
}

\usepackage{bibentry}

\begin{document}

\maketitle
\begin{abstract}
Large Multimodal Models (LMMs) often face a modality representation gap during pretraining: while language embeddings remain stable, visual representations are highly sensitive to contextual noise (e.g., background clutter). To address this issue, we introduce a visual comprehension stage, which we call \textbf{ViCToR}~(\textbf{Vi}sual \textbf{C}omprehension via \textbf{To}ken \textbf{R}econstruction), a novel pretraining framework for LMMs. 
ViCToR employs a learnable visual token pool and utilizes the Hungarian matching algorithm to select semantically relevant tokens from this pool for visual token replacement. Furthermore, by integrating a visual token reconstruction loss with dense semantic supervision, ViCToR can learn tokens which retain high visual detail, thereby enhancing the large language model’s (LLM’s) understanding of visual information.
After pretraining on 3 million publicly accessible images and captions, \textbf{ViCToR} achieves state-of-the-art results, improving over LLaVA-NeXT-8B by $10.4\%$, $3.2\%$, and $7.2\%$ on the MMStar, SEED$^{I}$, and RealWorldQA benchmarks, respectively. Code is available at \url{https://github.com/deepglint/Victor}.
\end{abstract}

\vspace{-0.5cm}

\section{Introduction}
Large Language Models (LLMs) \citep{touvron2023llama, achiam2023gpt} have achieved remarkable success in text understanding and generation, driven by autoregressive Transformer architectures and the scalability afforded by large-scale data and compute. However, these models remain fundamentally text-centric, limiting their applicability in multimodal contexts. To overcome this limitation, Large Multimodal Models (LMMs) have emerged, incorporating vision encoders such as CLIP~\citep{radford2021learning} to transform images into token-like representations that LLMs can process. For instance, LLaVA~\citep{liu2023visualinstructiontuning, liu2024improvedbaselinesvisualinstruction} leverages GPT-4~\citep{achiam2023gpt} to generate high-quality multimodal instruction-following datasets. Other approaches~\citep{li2023blip, zhu2023minigpt, wang2023cogvlm} have introduced more advanced projection modules, such as Q-Formers and expert-based mechanisms, to improve vision-language alignment and task performance.

\begin{figure}[!t]
\centering
\includegraphics[width=1.0\linewidth]{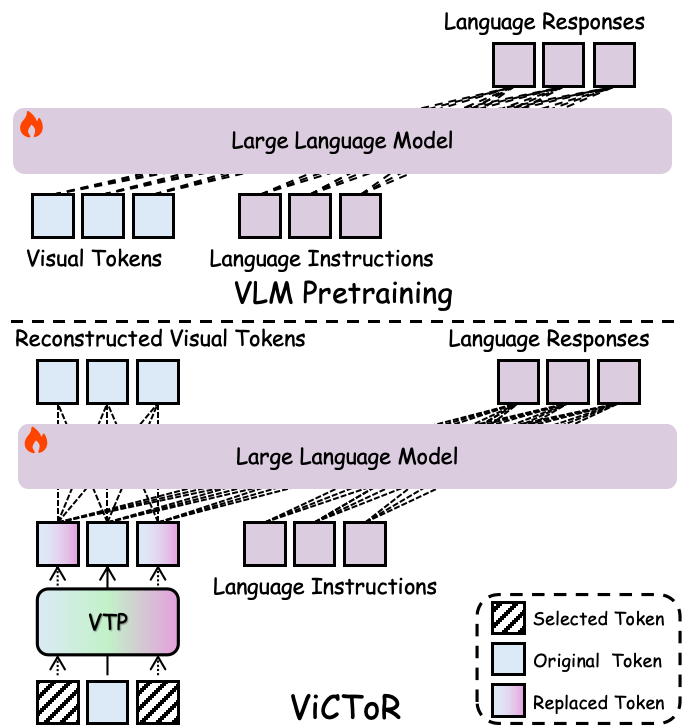}
\vspace{-4mm}
\caption{Traditional VLMs train language models to recognize visual tokens, while ViCToR instead replaces vision tokens with ones from a visual token pool, helping LLMs better understand and summarize images.}
\label{llava_attention}
\vspace{-6mm}
\end{figure}

Notably, the VILA framework~\citep{lin2024vila} pretrains with large-scale image-text interleaved data. However, these approaches suffer from relatively low efficiency. Recent insights suggest that using a relatively small amount of high-quality image captioning data, such as those from ShareGPT4v~\citep{chen2024sharegpt4v}, can lead to improved performance with a reduced computational cost. Another line of research~\citep{jin2023unified} introduces novel visual tokenizers that convert non-linguistic images into discrete token sequences, enabling LLMs to interpret them as foreign languages. Nevertheless, under such a unified framework, the intrinsic modality gap between vision and language remains unresolved, limiting the effectiveness of this approach.
 
To address these challenges, we introduce a visual comprehension stage and present \textbf{ViCToR}, a novel pretraining framework for LMMs. 
Specifically, we develop a learnable visual token pool (VTP) and adopt the Hungarian matching algorithm to select semantically relevant tokens from this pool for visual token replacement, as shown in Fig.~\ref{llava_attention}.
However, discretizing visual features into a limited set of tokens, while facilitating alignment with language, leads to severe loss of visual detail. To this end, we introduce a visual token reconstruction loss to maintain a faithful and effective visual representation.
After pretraining on 3 million publicly accessible images and captions, \textbf{ViCToR} achieves state-of-the-art performance on multiple downstream benchmarks. In summary, our \textbf{contributions} are the following:
\begin{itemize}[topsep=0pt, itemsep=0pt, parsep=0pt, partopsep=0pt]
\item We \textbf{design} a learnable visual token pool and employ the Hungarian algorithm to substitute selected original image tokens with the closest tokens from this pool.
\item We \textbf{propose} a visual token reconstruction stage using a reconstruction loss and dense semantic supervision.
\item We \textbf{demonstrate} that ViCToR achieves state-of-the-art performance on various benchmarks, surpassing LLaVA-NeXT-8B by up to $5.8\%$, showcasing strong capabilities in visual understanding and reasoning.
\end{itemize}

\section{Related Work}

\noindent{\bf Large Multimodal Model Pre-training.} LLaVA uses a subset of the CC3M~\citep{changpinyo2021cc12m} dataset for a more balanced coverage of concepts. 
Both the visual encoder and LLM are then frozen while the projection layer is trained to align the visual features and language tokens. 
However, relying solely on this approach will lead to a limited deep feature integration between the visual encoder and the LLM, mainly due to the restrictions imposed by the projection layer. 
To address this limitation, CogVLM~\citep{wang2023cogvlm} introduce a trainable visual expert module into the attention and feed-forward network layers of the language model. 
Despite this additional module, the LLM still remains limited by its frozen state and will continue to struggle interpreting the visual tokens.
LaVIT~\citep{jin2023unified} introduced a new visual tokenizer to convert images into a sequence of discrete tokens. However, directly inputting visual tokens into an LLM to enhance visual understanding with next-token prediction still presents significant difficulties. 
VILA~\citep{lin2024vila} proposes an interleaved pretraining stage to augment the LLM to support visual input, but it relies on a 50M pretraining dataset, requiring considerable computational resources. 
%
Recent studies such as ShareGPT4V~\citep{chen2024sharegpt4v} and LLaVA-OneVision~\citep{li2024llavaonevisioneasyvisualtask} demonstrate that high-quality image-caption pair data significantly improves the alignment between visual and textual modalities, thereby enabling more effective multimodal pretraining.

\noindent{\bf Visual Token Reconstruction.} Masked Image Modeling~(MIM)~\cite{he2021maskedautoencodersscalablevision,hondru2024maskedimagemodelingsurvey} is now a common pre-training strategy for improving visual comprehension.
Both 4M~\citep{mizrahi20234mmassivelymultimodalmasked} and MVP~\citep{wei2022mvp} propose to integrate this idea in the context of multimodal learning.
In contrast, MILAN~\citep{hou2022milan} proposes to reconstruct the image features infused with semantic content derived from caption supervision. 
Unmasked Teacher~\citep{li2023unmasked} selectively masks video tokens exhibiting low semantic content and aligns the remaining unmasked tokens through a linear projection to their counterparts from the teacher model. 
In a recent study, RILS~\citep{yang2023rils} introduced a novel pre-training framework that employs masked visual reconstruction within a language semantic space. This framework facilitates the extraction of structured information by vision models through the accurate semantic prediction of masked tokens. 
Meanwhile, EVA~\citep{fang2023eva} demonstrates that reconstructing the masked tokenized semantic vision features is an efficient strategy for vision-centric representation learning, removing the need for semantic feature quantization or any additional tokenization steps.
There are also several important works, such as Ross~\citep{wang2024reconstructivevisualinstructiontuning} and Show-O~\citep{xie2024showosingletransformerunify}, that leverage visual reconstruction tasks to train large multimodal models (LMMs). These approaches utilize large language models (LLMs) to reconstruct visual features, thereby enhancing both visual understanding and generation capabilities.
Inspired by the above works, we propose a visual token reconstruction task for pretraining LMMs to bridge the modality gap between LLMs and visual tokens, enhancing the models’ understanding of visual information. 
\vspace{-0.5cm}

\section{Method}
\vspace{-0.3cm}
\textbf{Preliminaries.} Given an input image $I \in \mathbb{R}^{H \times W \times C}$, the vision encoder divides it into $N$ patches $p_i$, each transformed into an embedding $z_i = E_v(p_i) \in \mathbb{R}^D$, where $D$ represents the embedding dimension. This process forms a sequence $z_i$. These visual tokens are continuous and encode localized spatial information of the image. In contrast, after passing text through its tokenizer and embedding layer, each token $t_i$ in a sequence $\{t_i\}_{i=1}^L$ will be mapped via an embedding matrix $E \in \mathbb{R}^{|V| \times D}$, where $V$ denotes the size of the discrete vocabulary, resulting in embeddings $e_i = E(t_i)$. These tokens are discrete and globally shared, with their relationships derived from co-occurrence statistics in large text corpora. This fundamental difference leads to the modality representation gap. Despite existing efforts such as LLaVA~\citep{liu2024llava}, which attempts to align the visual encoder and LLMs into a common feature space using a learnable mapping, this mapping does not align with the LLM’s original training paradigm, which is designed for discrete, symbolic language tokens. As a result, the model is only passively exposed to image features in a transformed, language-like form, making it difficult for the LLM to fully understand and internalize the representational structure and inductive biases of visual data.

\noindent \textbf{ViCToR.} To overcome the limitations of existing pretraining paradigms, we propose a novel framework, ViCToR, for the pre-training of LMMs. In Sec.~\nameref{sec:token_pool}, we introduce the VTP for discretizing visual features into a limited set of tokens. In Sec.~\nameref{visual_token_gen}, we then propose a visual token reconstruction task to recover the loss in visual detail from the token pool. Finally, in Sec.~\nameref{sec:detailed_caption}, we show how to utilize the detailed captions to provide dense semantic supervision for vision token reconstruction. The complete training pipeline is outlined in Sec.~\nameref{training_pipeline}.

\begin{figure*}[t!]
\centering
\includegraphics[width=0.95\linewidth]{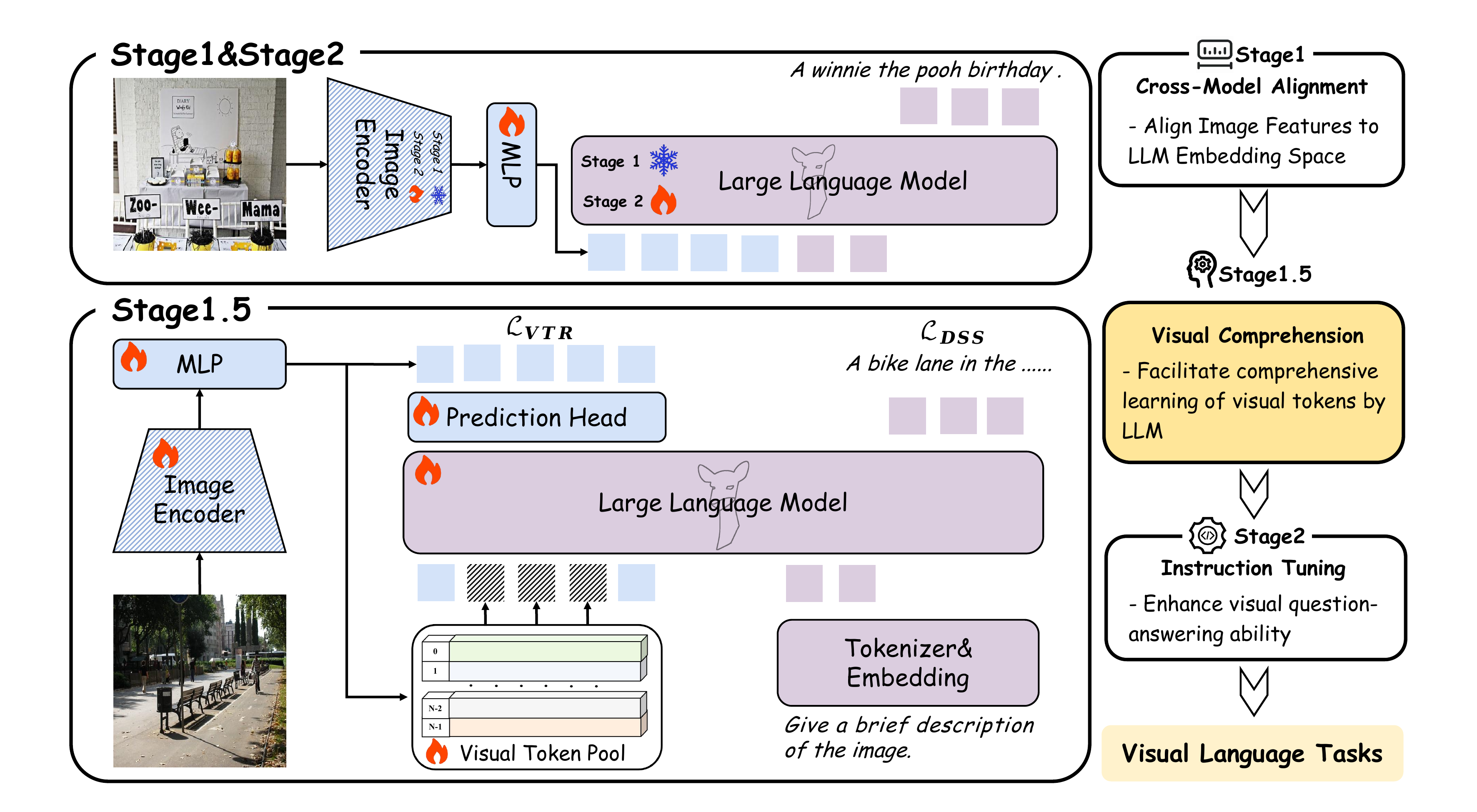}
\vspace{-3mm}
\caption{The training pipeline of our proposed ViCToR model. In contrast to LLaVA-1.5~\citep{liu2024llava_improved}, we introduce an additional pre-training stage that involves visual token reconstruction and dense semantic supervision. This stage is essential for improving visual comprehension.}
\vspace{-6mm}
\label{fig:training_pipeline}
\end{figure*}

\subsection{Visual Token Pool}
\label{sec:token_pool}

To bridge the gap between continuous visual tokens and discrete language tokens, we introduce a VTP consisting of reusable and learnable visual tokens. Each selected token in a sequence of visual tokens is replaced with a learnable token selected from this pool. These tokens capture semantically meaningful visual patterns common across multiple samples, thereby enabling the LLM to actively learn the mapping between continuous tokens from the vision encoder and discrete tokens from the vision token pool within the language embedding space. We denote the VTP as $T_{p} \in \mathbb{R}^{N\times D}$, where $N$ and $D$ represent the number of learnable visual tokens and the feature dimension respectively. With a selection ratio of $\gamma$, we obtain the set of selected visual tokens $\widetilde{T}_{v}$. The selected tokens $\widetilde{T}_{v}$ are first padded with $\varnothing$ to maintain a consistent set size of $N$ prior to assignment. \\
\noindent{\bf Token Assignment.} We investigate two distinct methods for assigning tokens from the token pool. The first approach involves a simple nearest-neighbor~\citep{1053964} lookup. However, this method often results in an over-reliance on a limited subset of tokens, leading to suboptimal utilization of the entire token space (refer to Sec.~\nameref{sec:ablation}). Fortunately, this assignment issue is extensively addressed in combinatorics. The Hungarian algorithm~\cite{kuhn1955hungarian} provides a polynomial-time solution, with numerous efficient GPU adaptations available~\cite{hungarian_papad, rectangular_assignment}. Owing to its versatility, the algorithm has been widely adopted in various domains, including object detection~\cite{detr}.

Our objective is to determine a bipartite matching between $\widetilde{T}{v}$ and $T{p}$ that minimizes the total cost, defined here as the L2 distance between the replaced and original visual tokens. We address this by identifying a permutation of $N$ elements, $\sigma \in \mathfrak{S}_N$, that minimizes this cost metric:\\
\\
\vspace{-2mm}
\begin{equation}
    \hat{\sigma} = \underset{\sigma \in \mathfrak{S}_N}{\arg \min } \sum_i^N \| \widetilde{T}_{v}^{i} - T_{p}^{\sigma(i)} \|_2
\end{equation} \\ 
The Hungarian algorithm addresses this assignment by iteratively subtracting the minimum values from each row and column. We observe that this optimization step is computationally inexpensive, constituting less than 5\% of the total wall-clock time of the forward pass during training.

\begin{table*}[t!]
  \centering
  \scriptsize
  \setlength{\tabcolsep}{3pt}
  \resizebox{1.0\textwidth}{!}{%
    \begin{tabular}{llc|ccccccccc}
      \toprule
      \textbf{Method} & \textbf{Pub.} & \textbf{Res.}  & \textbf{MMStar} & \textbf{RealWorldQA} & \textbf{MMBench$^{en}_{val}$} & \textbf{OCRBench}  & \textbf{POPE} & \textbf{MMMU} & \textbf{AI2D} & \textbf{MME}  & \textbf{SEED$^{I}$} \\
      \midrule
      
      LLaVA-1.5-13B  & NeurIPS'23 & 336$^2$   & 34.3       & 55.3       &67.8       & 337      & \textbf{88.4}       & 37.0       & 61.1    & 1781 & 68.2  \\

      LLaVA-NeXT-8B  & CVPR'24 & 672$^2$   & 43.9       & 58.4       & --       & 531        & 87.1       & 43.1       & 72.8    & \underline{1908}  & 72.5 \\
      Cambrian-13B  & NeurIPS'24 & 1024$^2$   & 47.1       & \underline{63.0}       & \underline{75.7}       & \underline{610}        & 86.8       & 41.6       & 73.6    & 1877  & \underline{74.4} \\
      IDEFICS2-8B  & NeurIPS'24 & 768$^2$   & 49.5       & 60.7       & --       & \textbf{626}        & 86.2       & 45.2       & 72.3    & 1848  & 71.9 \\
      Mantis-8B & TMLR'24  & 384$^2$   & 41.3       & 52.2       & --       & 347        & 84.0       & 41.1       & 60.4    & 1675  & 68.5 \\
      Ross & ICLR'25 & 384$^2$   & \underline{53.9}       & 58.7       & --       & 553            & \underline{88.1}       & \textbf{49.0}       & \underline{79.4}      & 1854 & 73.6      \\
      \midrule
      ViCToR-7B  &    & 384$^2$   & \textbf{54.3}    & \textbf{65.6} & \textbf{79.0} & 556  & \textbf{88.4}       & \underline{48.9}      & \textbf{79.5}     & \textbf{2071} & \textbf{75.7}  \\
      \bottomrule
    \end{tabular}%
  }
  \vspace{-2mm}
  \caption{\small Comparison with other state-of-the-art vision-language models on various VLM benchmarks demonstrates that our method achieves leading performance across multiple domains. We highlight the best results in \textbf{bold} and the second-best results with an \underline{underline}. All results of other methods reported in the tables are taken from their official papers and the Open VLM Leaderboard~\citep{duan2024vlmevalkit}.}
  \label{tab:academic}
  \vspace{-2mm}
\end{table*}
\begin{table*}[t!]
  \centering
  \scriptsize
  \setlength{\tabcolsep}{3pt}
  \resizebox{1.0\textwidth}{!}{%
    \begin{tabular}{l c c c c c c | c c c c }
      \toprule
      \textbf{Method} & \textbf{Pub.} & \textbf{VE} & \textbf{LLM} & \textbf{Res.} & \textbf{Pretrain} & \textbf{Finetune} & 
      \textbf{AI2D} & \textbf{MME}  & \textbf{MMStar} & \textbf{RealWorldQA}   \\
      \midrule
      
      LLaVA-NeXT-7B & CVPR'24 & CLIP-L/14 & Vicuna1.5-7B  & 672$^2$ & 558K        & 780k     & 67.0         & 1769        & 37.6     & 57.8            \\
      VILA1.5 & CVPR'24 & SigLIP-400M/14  & LLaMA3-8B         & 384$^2$ & 50m          & 1m     & 58.8       &  1648       & 39.7     & 43.4           \\
      Sharegpt4V & ECCV'25 & CLIP-L/14 & Vicuna1.5-7B    & 336$^2$ & 558K+1.2m   & 742k     & 58.0        & \textbf{1915}      & 35.7     & 54.9             \\
      \midrule
      ViCToR-7B   &   & CLIP-L/14 & Vicuna1.5-7B & 336$^2$ & 558K+1.2m   & 780k     & \textbf{70.9}   & 1873      & \textbf{41.4}     & \textbf{58.3}     \\
      \bottomrule
    \end{tabular}%
  }
  \vspace{-2mm}
  \caption{\small{Comparison with other advanced pretraining methods for a fair evaluation. We mark the best performance \textbf{bold}. All results of other methods reported in the tables are taken from their official papers and the Open VLM Leaderboard~\citep{duan2024vlmevalkit}.}}
  \label{tab:compare}
  \vspace{-6mm}
\end{table*}

\subsection{Visual Token Reconstruction}
\label{visual_token_gen}
For this reconstruction task, we follow the same training paradigm of LLaVA, where an input image $I$ is first encoded into a sequence of visual tokens. Specifically, visual features are extracted using a pretrained vision encoder (e.g., CLIP or SigLip2) $E_v$, and then projected into the language embedding space through a Multi-Layer Perceptron (MLP):
\[
T_v = \{v_1, v_2, \dots, v_n\} = \mathrm{MLP}(E_v(I)) \in \mathbb{R}^{n \times d},
\]
where $T_v$ denotes the sequence of $n$ visual tokens, each of dimension $d$.
For the visual reconstruction task, we randomly select a proportion $\gamma$ of the visual tokens from $T_v$. Let $\mathcal{M} \subset \{1, 2, \dots, n\}$ indicate the index set of the selected tokens, sampled uniformly at random such that $|\mathcal{M}| \approx \gamma n$. The selected token set is present as $\widetilde{T}_v = \{v_i \mid i \in \mathcal{M} \}$ and these tokens are replaced by visual tokens drawn from a VTP, resulting in a new visual input $\hat{T}_v$.

To supervise the reconstruction objective, for each index $i\in\mathcal{M}$, the model predicts a reconstructed visual token $\hat{v}_i$. These predictions form the reconstructed visual token set $\hat{T}_v = \{\hat{v}_i\mid i\in\mathcal{M}\}$.
The reconstruction loss is then computed by measuring the distance between the predicted visual tokens $\hat{v}_i$ and the grounded visual tokens $v_i$ for all selected positions. The loss of visual token reconstruction is defined as
\[
\mathcal{L}_\mathrm{VTR} = \sum_{i \in \mathcal{M}} \| \hat{v}_i - v_i \|.
\]

\subsection{Dense Image Captioning Task}
\label{sec:detailed_caption}
Captions which provide detailed visual semantic information can facilitate high-quality visual token reconstruction in LLMs. Thus, we introduce an additional task of detailed caption generation to help the model establish stronger associations between visual and language modalities.

Since visual token reconstruction encourages the LLM to actively model visual information by providing dense supervision, it is naturally complemented by a detailed caption generation task. This synergy helps the model form stronger associations between visual and linguistic modalities, leading to better generation capabilities. 
We minimize the loss of dense semantic supervision (DSS) by optimizing the negative log-likelihood of the target text tokens \( T_{rt} \), conditioned on the visual context \( \hat{T}_v \) and the previously generated tokens:
\begin{equation}
\mathcal{L}_{\mathrm{DSS}} = -\sum_i \log p\left(t_i \mid \hat{T}_v, t_1, \cdots, t_{i-1} \right).
\end{equation}

Finally, the overall loss function is defined as:
\begin{equation}
\mathcal{L} = \alpha \mathcal{L}_{\mathrm{VTR}} + \mathcal{L}_{\mathrm{DSS}},
\end{equation}
where $\alpha$ is a loss weight to balance the influence of different losses.

\begin{figure*}[t]
\centering
\includegraphics[width=0.9\linewidth]{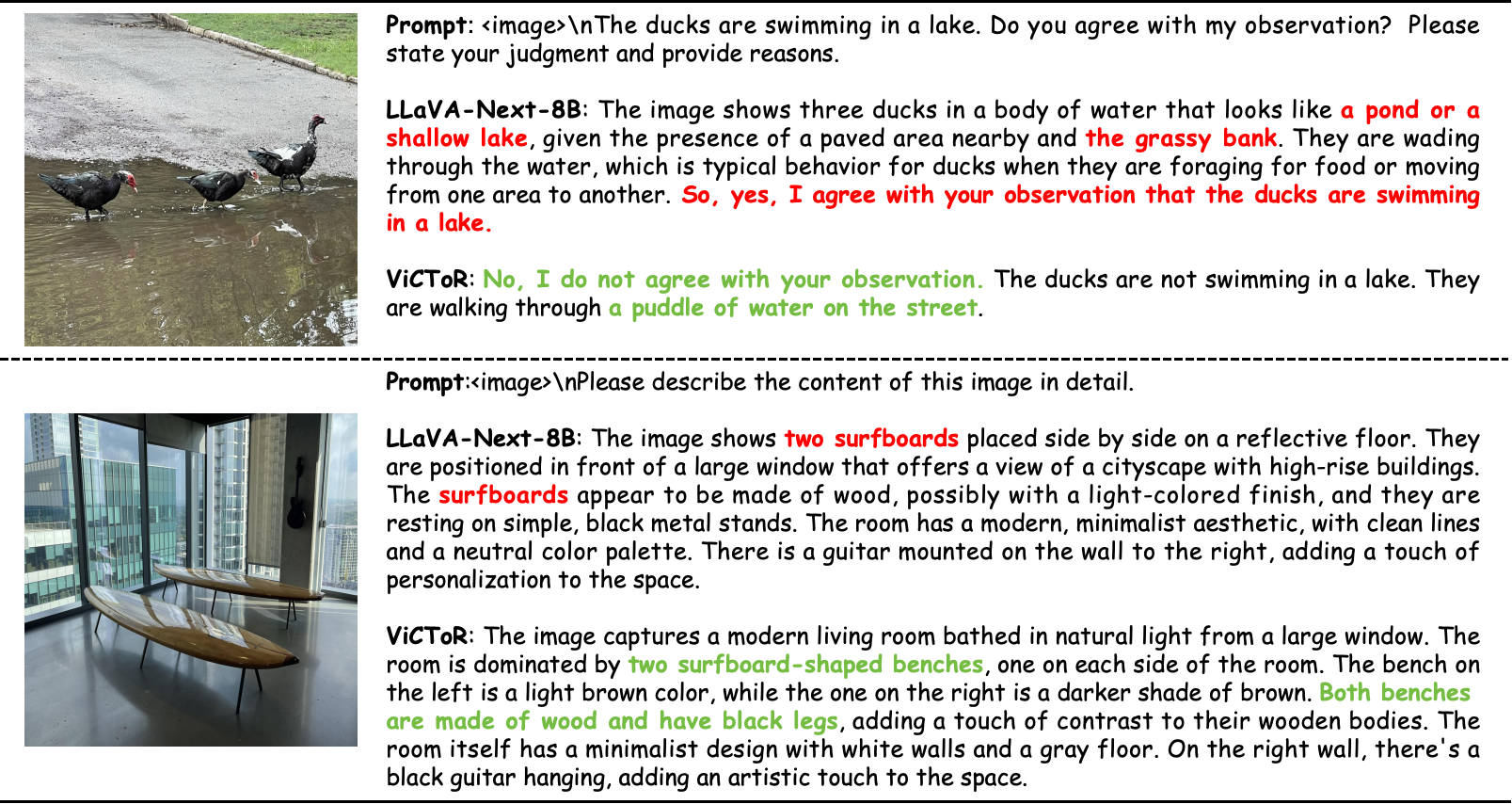}
\caption{\small{Qualitative Comparison of LLaVA-Next-8B and ViCToR-7B. Benefiting from our proposed cross-modal comprehension stage, the ViCToR model exhibits enhanced visual comprehension and reasoning capabilities. Moreover, it can generate enriched image descriptions.}}
\label{fig:case}
\vspace{-4mm}
\end{figure*}

\subsection{Training Pipeline}
\label{training_pipeline}
The overall training pipeline of our ViCToR model is illustrated in Fig.~\ref{fig:training_pipeline}. Building upon LLaVA-1.5, we design a comprehensive three-stage training strategy, consisting of two pre-training stages followed by instruction tuning.

\noindent{\bf Stage 1: Cross-modal Alignment.} 
Following the LLaVA-1.5, we initially pretrain the projection layer using the identical 558K dataset employed by LLaVA-1.5. This stage aims to align image features with the LLM embedding space. During this process, we freeze the weights of both the visual encoder and the LLM, allowing exclusive updates to the projection layer.

\noindent{\bf Stage 1.5: Visual Comprehension.}  
To further enhance the model’s cross-modal understanding, we introduce an intermediate cross-modal reconstruction phase after alignment. In this stage, we pretrain the entire model, encompassing the visual encoder, projection layer, and LLM, utilizing three million high-quality image-text pairs from the LLaVA-ReCap-CC3M dataset~\citep{li2024llavaonevisioneasyvisualtask}.

\noindent{\bf Stage 2: Instruction Tuning.}  
Finally, we perform instruction tuning on the whole model using 780K instruction-following samples from the LLaVA-NeXT~\citep{liu2023improvedllava}. This stage further improves the model’s ability to follow human instructions and performing a variety of multimodal tasks.

\section{Experiments}

\textbf{Implementation Details. } 
Following LLaVA-1.5, we explore different model configurations for our framework. For main experiments (Tab. \ref{tab:academic}), we employ the SigLIP2 So400m/14@384px~\citep{tschannen2025siglip2multilingualvisionlanguage} as visual encoder and utilize Qwen2.5-7B ~\citep{qwen2025qwen25technicalreport} as the large language model (LLM). 
To further demonstrate the effectiveness of our method, we also follow the LLaVA-1.5 setup by using CLIP ViT-L/14@336px~\citep{radford2021learning} as the visual encoder and Vicuna~\citep{vicuna2023} 7B as the LLM (Tab. \ref{tab:compare}.)

During the cross-modal alignment stage, we set the projection layer learning rate to 1e-3. For visual comprehension stage, learning rates are set to 2e-5 for both the LLM , projection layer and VTP, 2e-6 for the vision encoder. We use a loss weight $\alpha$ of 10, random replace ratio $\gamma$ of $75\%$, and the VTP size of $2048$. In instruction tuning, learning rates are 2e-5 for the LLM and projection layer, and 2e-6 for the vision encoder. AdamW~\citep{loshchilov2017decoupled} is used as the optimizer with weight decay $0.2$ and $\beta_{1}$/$\beta_{2}$ of $0.9$/$0.98$. ViCToR is trained on $16 \times$ NVIDIA A800 for 35 hours.\\

\noindent \textbf{Evaluation Benchmarks.} We evaluate ViCToR across various benchmarks, including 1) OCR-Related Question Answering: OCRBench~\citep{Liu_2024}; 
2) hallucination: POPE~\citep{li2023evaluating}; 3) Comprehensive Reasoning Benchmarks: MMBench~\citep{liu2023mmbench}, 
RealWorldQA~\citep{grok1.5v}, MMStar~\citep{chen2024we}, MME~\citep{fu2024mmecomprehensiveevaluationbenchmark} and SeedBench$^{I}$~\citep{li2023seedbenchbenchmarkingmultimodalllms}; 4) Science Visual Question Answering: MMMU~\citep{yue2024mmmumassivemultidisciplinemultimodal} and AI2D~\citep{kembhavi2016diagram}.

\begin{figure}[!t]
\centering
\includegraphics[width=\linewidth]{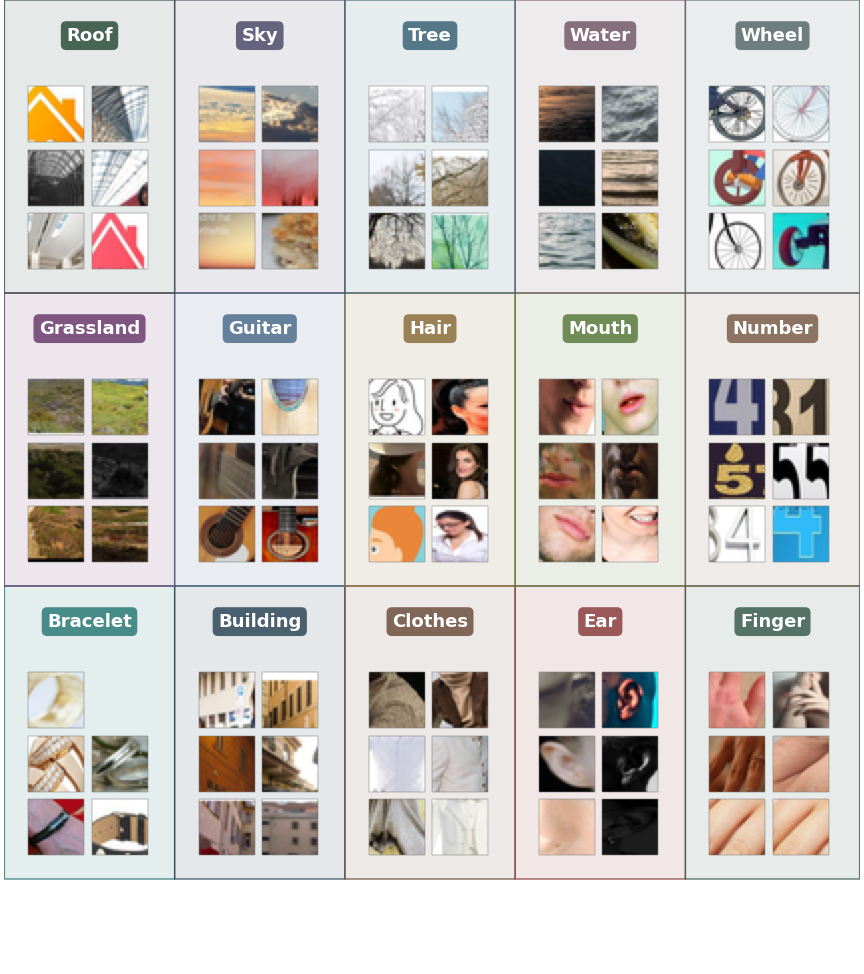}
\vspace{-15mm}
\caption{\small{We select and visualize image regions consisting of more than four contiguous local patches that exhibit the shortest distance to the same item in the VTP.}}
\label{fig:vtp}
\vspace{-6mm}
\end{figure}

\begin{figure*}[t!]
\centering
\includegraphics[width=\linewidth]{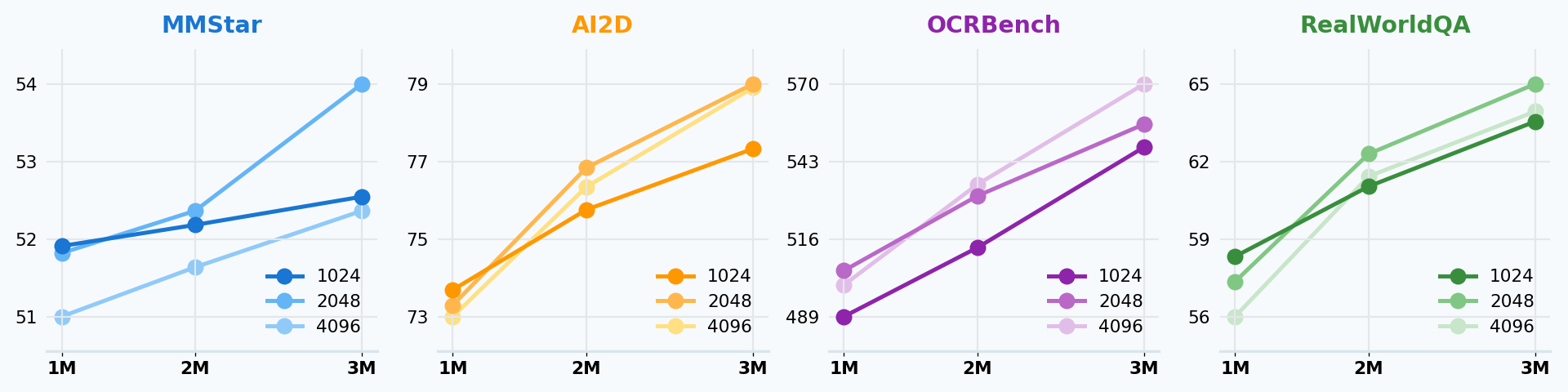}
\vspace{-5mm}
\caption{\small{Performance comparison of different pre-training data scales and token pool sizes across various benchmark types.}}
\label{fig:pool_data_size}
\vspace{-5mm}
\end{figure*}

\subsection{Main Results}
\vspace{-1mm}

Tab.~\ref{tab:academic} presents a comprehensive comparison of our model with the LLaVA series, Cambrian~\cite{tong2024cambrian1fullyopenvisioncentric}, IDEFICS2~\cite{laurençon2024mattersbuildingvisionlanguagemodels}, Mantis~\cite{jiang2024mantisinterleavedmultiimageinstruction}, and Ross across multiple benchmarks. Although other methods typically utilize higher input resolutions or larger training datasets, our model consistently achieves superior results on these benchmarks. Moreover, by using less training data, our approach not only significantly reduces training costs but also lowers inference costs due to the reduced input resolution. In addition, Fig.~\ref{fig:case} showcases a number of examples highlighting the concrete differences in capabilities between our model and LLaVA-NeXT-8B.

\noindent{\bf Fair Comparative Experimental Designs.} 
To provide an intuitive and rigorous evaluation of the effectiveness of our pre-training approach, we conduct comprehensive comparisons with other mainstream pre-training methods, including VILA and ShareGPT4V, under fair or even slightly disadvantageous conditions for our method. Specifically, we sample 1.2M pre-training examples, employ CLIP-L/14@336 as the vision encoder, and use Vicuna-1.5-7B as the language model. For baseline comparison, we select LLaVA-NeXT-7B with an image tiling strategy and ensure that its actual number of training tokens is at least comparable to ours. The final results, as shown in Tab.~\ref{tab:compare}, demonstrate the comparative performance of our method.

\begin{table}
  \centering
  \scriptsize
  \setlength{\tabcolsep}{3pt}
  \resizebox{0.48\textwidth}{!}{%
    \begin{tabular}{l|cccc}
      \toprule
      \textbf{Method} & \textbf{MMStar} & \textbf{AI2D} & \textbf{OCRBench} & \textbf{RealWorldQA}  \\
      \midrule
      ViCToR$_{vq-gan}$  & 50.3       & 76.1       & 535       & 62.8        \\
      ViCToR$_{kmeans}$ & 52.8       & 76.3       & 548       & 61.3  \\
      \midrule
      ViCToR           & \textbf{53.5}       & \textbf{79.8}       & \textbf{564}       & \textbf{64.4} \\
      \bottomrule
    \end{tabular}%
  }
  \vspace{-2mm}
  \caption{\small Comparison of different VTP initialization strategies on model performance across various benchmarks.}
  \label{tab:vtp_init}
  \vspace{-6mm}
\end{table}

\vspace{-2mm} 
\subsection{Insights and Analysis of the Visual Token Pool}
\vspace{-1mm}
We propose to use a VTP for bridging the modality gap between visual and textual tokens. We believe that this VTP enables the large language model (LLM) to easily aggregate region-level image features from all input images observed during training. To evaluate the classification capability of VTP-regional image features after pre-training, we group and visualize image patches that are closes to the same item in the VTP (see Fig.~\ref{fig:vtp}). The results clearly demonstrate that VTP can effectively cluster objects of the same category, even when they exhibit various visual forms, such as ``wheel'' and ``roof''. 

\begin{table*}[t]
  \centering
  \scriptsize
  \setlength{\tabcolsep}{4pt}
  \renewcommand{\arraystretch}{1.15}
  \subfloat[\small Effect of training objective.\label{table:attention}]{
    \begin{minipage}{0.48\linewidth}
      \centering
      \resizebox{\textwidth}{!}{
        \begin{tabular}{l|ccccc}
          \toprule
          \textbf{Method} & \textbf{MME} & \textbf{OCRBench} & \textbf{POPE} & \textbf{SEED$^{I}$} & \textbf{RealWorldQA} \\
          \midrule
          ViCToR$_{pixel}$      & 1764 & 447 & 86.3 & 69.3 & 59.7 \\
          ViCToR$_{QFormer}$    & 1865 & 512 & 87.5 & 72.1 & 61.5 \\
          ViCToR            & \textbf{2071} & \textbf{556} & \textbf{88.4} & \textbf{75.7} & \textbf{65.6} \\
          \bottomrule
        \end{tabular}
      }
    \end{minipage}
  }
  \hfill
  \subfloat[\small Effect of replace ratio.\label{table:replace_ratio}]{
  \renewcommand\arraystretch{1.1}
    \begin{minipage}{0.48\linewidth}
      \centering
      \resizebox{\textwidth}{!}{
        \begin{tabular}{c|ccccc}
          \toprule
          \textbf{Replace Ratio} & \textbf{MME} & \textbf{OCRBench} & \textbf{POPE} & \textbf{SEED$^{I}$} & \textbf{RealWorldQA} \\
          \midrule
          0.50      & 1908          & 532           & \textbf{88.5} & 74.2          & 60.1 \\
          
          0.75      & \textbf{2071} & \textbf{556}  & 88.4      & \textbf{75.7} & \textbf{65.6} \\
          
          0.90      & 1978          & 528           & 88.0          & 74.8          & 63.4 \\
          \bottomrule
        \end{tabular}
      }
    \end{minipage}
  }
  \hfill
  \\
  \subfloat[\small Effect of matching algorithm.\label{table:matching}]{
    \begin{minipage}{0.48\linewidth}
      \centering
      \resizebox{\textwidth}{!}{
        \begin{tabular}{l|ccccc}
          \toprule
          \textbf{Method} & \textbf{MME} & \textbf{OCRBench} & \textbf{POPE} & \textbf{SEED$^{I}$} & \textbf{RealWorldQA} \\
          \midrule
          Nearest Neighbors  & 1919          & 532          & 87.2          & 72.2          & 63.2 \\
          Hungarian Matching & \textbf{2071} & \textbf{556} & \textbf{88.4} & \textbf{75.7} & \textbf{65.6} \\
          \bottomrule
        \end{tabular}
      }
    \end{minipage}
  }
  \hfill
  \subfloat[\small Effect of removing Stage 1.\label{table:stage1}]{
  \renewcommand\arraystretch{0.95}
    \begin{minipage}{0.48\linewidth}
      \centering
      \resizebox{\textwidth}{!}{
        \begin{tabular}{cc|ccccc}
          \toprule
          \textbf{S1} & \textbf{S1.5} & \textbf{MME} & \textbf{OCRBench} & \textbf{POPE} & \textbf{SEED$^{I}$} & \textbf{RealWorldQA} \\
          \midrule
          \xmark & \cmark & 1884          & 514          & 86.3          & 72.8             & 63.1 \\
          \cmark & \cmark & \textbf{2071} & \textbf{556} & \textbf{88.4} & \textbf{75.7} & \textbf{65.6} \\
          \bottomrule
        \end{tabular}
      }
    \end{minipage}
  }
  
  \vspace{-3mm}
  \caption{\small Results of ablation studies on ViCToR-7B, evaluating the effects of different training objectives, replace ratios, matching algorithms, and the presence of Stage 1 across multiple benchmarks.}
  \label{table:ablation}
  \vspace{-2mm}
\end{table*}

\noindent{\bf What is the most effective initialization strategy for the VTP?} In our approach, we introduce VTP to bridge the modality gap between linguistic and visual tokens, and we adopt random initialization. During training, the VTP is adaptively learned in an end-to-end manner under the joint supervision of vision and language signals by the LLM. This raises a natural question: does initializing the VTP with visual features prior to training further enhance the LLM’s understanding of visual representations?
To investigate this, we employ two approaches to pre-initialize the VTP using visual features extracted from 3M images: VQ-GAN~\citep{esser2021tamingtransformershighresolutionimage} reconstruction and K-means clustering, resulting in \textbf{ViCToR$_{vq-gan}$} and \textbf{ViCToR$_{kmeans}$}. We compare these strategies with the standard random initialization, and the results are reported in Tab.~\ref{tab:vtp_init}.

The results show that initializing the VTP with visual features does not lead to performance gains and, in some cases, results in performance degradation. This finding supports our hypothesis that the optimal VTP representation should be adaptively constructed by the LLM based on its integrated understanding of both vision and language, rather than simply inheriting the feature distribution from the visual encoder.

To systematically investigate the relationship between the size of the VTP and the amount of pre-training data, we conduct experiments across four benchmark datasets. Specifically, we consider three scales of pre-training data and evaluate model performance with VTP sizes of 1024, 2048, and 4096 for each data scale. As shown in Fig.~\ref{fig:pool_data_size}, increasing the VTP size leads to significant performance gains only when sufficient pre-training data is available. These results highlight the necessity of jointly scaling both VTP size and pre-training data to fully realize the model’s potential.

\subsection{Ablation Study}
\label{sec:ablation}

\begin{figure}[!t]
\centering
\includegraphics[width=0.95\linewidth]{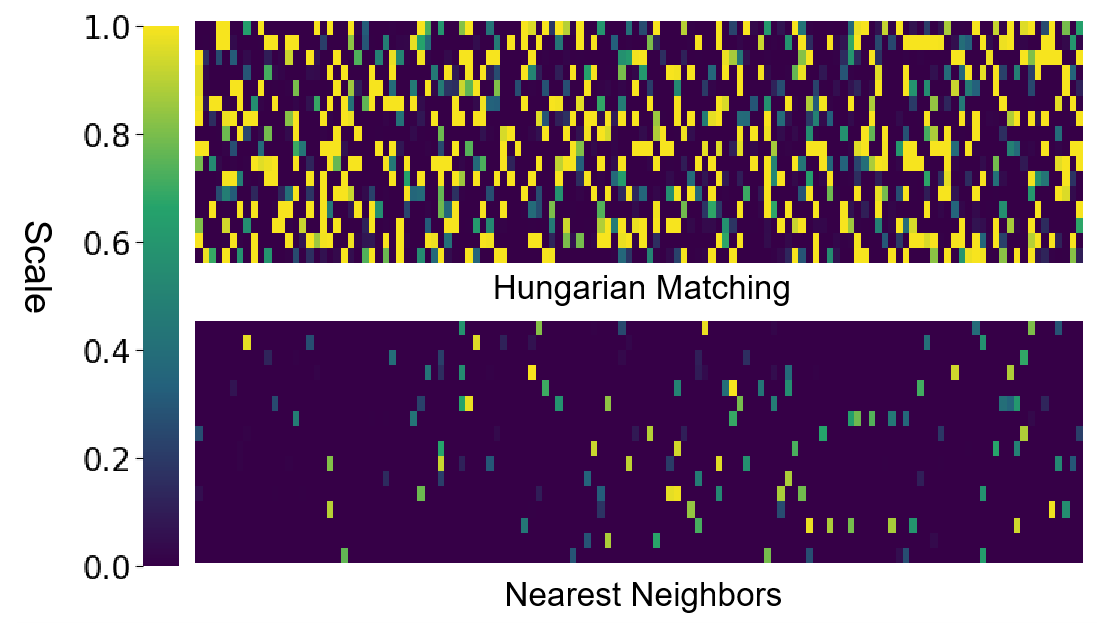}
\vspace{-3mm}
\caption{\small{The token utilization rate in VTP with different matching algorithms with the ViCToR-7B model.}}
\label{fig:utilization}
\vspace{-6mm}
\end{figure}

\noindent{\bf Different reconstruction objectives and model architectures.} 
In ViCToR, we use LLMs to reconstruct visual features for image understanding. We also test reconstructing pixels instead of features, keeping the MAE decoder and loss~\citep{he2022masked} and replacing the MLP adapter with QFormer~\cite{li2023blip2bootstrappinglanguageimagepretraining}. As Tab.~\ref{table:attention} shows, pixel reconstruction, despite its low training loss, leads to worse downstream performance, suggesting that over‐reliance on the decoder hinders efficient LLM comprehension of visual tokens. QFormer likewise offers no clear benefit.

\begin{table*}[t!]
\centering
\renewcommand{\arraystretch}{1.15}
\setlength{\tabcolsep}{4pt}
\resizebox{0.8\linewidth}{!}{
\scriptsize
\begin{tabular}{ll|ccccccccc}
\toprule
 $\mathcal{L}_{VTR}$ & $\mathcal{L}_{DSS}$ & \textbf{MMStar} & \textbf{RealWorldQA} & \textbf{MMBench$^{en}_{val}$} & \textbf{OCRBench} & \textbf{POPE} & \textbf{MMMU} & \textbf{AI2D} & \textbf{MME}  & \textbf{SEED$^{I}$} \\
\midrule
\Checkmark   & \XSolidBrush & 50.3 & 63.2 & 75.4 & 516 & \textbf{88.4} & 48.3 & 78.1   & 1915 & 73.2 \\
\XSolidBrush & \Checkmark   & 51.9 & 62.5 & 76.1 & 535 & 87.5 & 48.7 & 77.7   & 1847  & 73.9\\
\Checkmark   & \Checkmark   & \textbf{54.3}    & \textbf{65.6} & \textbf{79.0} & \textbf{556}  & \textbf{88.4}       & \textbf{48.9}      & \textbf{79.5}     & \textbf{2071} & \textbf{75.7}\\
\bottomrule
\end{tabular}
}
\vspace{-3mm}
\caption{\small Comparison of ablation results for $\mathcal{L}_{VTR}$ and $\mathcal{L}_{DSS}$ on multiple multimodal benchmarks.}
\label{table:ablation_loss}
\vspace{-5mm}
\end{table*}

\noindent{\bf Visual Token Replace Ratio.} 
The replace ratio of visual tokens directly affects the difficulty of the visual token reconstruction task, and thus the effectiveness of pretraining.
In Tab.~\ref{table:replace_ratio}, we report the results of experiments with different mask ratios. 
Similar to the observations with MAE, we find that using a 75\% replacement ratio yields optimal results in several downstream benchmarks. 
Lower ratios, e.g., 50\%, make the pre-training task too easy, while higher ratios, e.g., 90\%, make it too difficult.

\noindent{\bf Nearest Neighbors v.s. Hungarian Matching.} 
To improve the utilization rate of the tokens in the VTP, we use the Hungarian Matching. This is important to avoid an over-dependence on a small subset of the token pool, while also enabling a sufficient exploration of the VTP space. In Fig.~\ref{fig:utilization}, we present a comparative analysis of token utilization using both Nearest Neighbors and Hungarian Matching. Due to the Hungarian Matching requirement to select distinct visual tokens, we observe an improvement in the overall utilization of VTP, leading to a improvement across many evaluation benchmarks (Tab.~\ref{table:matching}). 

\noindent{\bf Stage 1 plays a critical role in our overall framework.}
Since the visual replace operation is performed after the visual features are processed by the adapter, Stage 1 is necessary to establish an initial alignment between the visual and linguistic spaces. Ablation studies (see Tab.~\ref{table:stage1}) demonstrate that omitting Stage 1 leads to a significant drop in performance and less efficient training, making it difficult for the model to achieve effective alignment.

\noindent{\bf Stage 1.5 Loss Ablation Study of ViCToR}
We conducted an ablation study on the two loss functions used in the Stage 1.5 phase of ViCoTR: the visual reconstruction loss($\mathcal{L}_{VTR}$) and the dense sequence supervision loss($\mathcal{L}_{DSS}$). Specifically, we systematically analyzed the effects of using each loss individually and in combination across nine different benchmarks, as shown in Tab.~\ref{table:ablation_loss}. The results demonstrate that $\mathcal{L}_{VTR}$ helps suppress visual hallucinations (e.g., on POPE) and excels in real-world scenario question answering tasks such as RealWorldQA, while $\mathcal{L}_{DSS}$ is more suitable for tasks requiring dense visual information understanding, such as OCR. Combining both losses achieves optimal performance on most benchmarks.

\section{Conclusion}
 We present ViCToR, a novel pretraining methodology that significantly enhances the visual comprehension capabilities of Large Multimodal Models (LMMs). Our approach introduces a visual comprehension stage that effectively bridges the visual and textual domains, coupled with a dynamically learnable VTP leveraging the Hungarian algorithm for precise visual semantic processing. Extensive comparative experiments demonstrate that ViCToR outperforms state-of-the-art methods across multiple benchmark datasets, with particularly strong performance in visual understanding and cross-modal reasoning tasks. Through comprehensive ablation studies, we validate the efficacy and necessity of each component, with particular evidence supporting the critical contribution of the dynamic VTP to the model's overall performance. This work establishes a new paradigm for enhancing visual comprehension in multimodal foundation models and lays a solid foundation for the advancement of vision-language models.

\section*{Limitations}

In this work, we have focused solely on image token reconstruction, which limits the scope to static images. However, for comprehensive video understanding, it is essential to consider both spatial and temporal token reconstruction. This would allow us to capture the dynamic changes that occur across frames and enhance the model’s ability to process and interpret video sequences more effectively. Expanding our approach to include spatial-temporal token reconstruction is a necessary step for future improvements in video analysis.

\bibliography{main}

\begin{thebibliography}{51}
\providecommand{\natexlab}[1]{#1}

\bibitem[{Achiam et~al.(2023)Achiam, Adler, Agarwal, Ahmad, Akkaya, Aleman, Almeida, Altenschmidt, Altman, Anadkat et~al.}]{achiam2023gpt}
Achiam, J.; Adler, S.; Agarwal, S.; Ahmad, L.; Akkaya, I.; Aleman, F.~L.; Almeida, D.; Altenschmidt, J.; Altman, S.; Anadkat, S.; et~al. 2023.
\newblock Gpt-4 technical report.
\newblock \emph{arXiv:2303.08774}.

\bibitem[{Carion et~al.(2020)Carion, Massa, Synnaeve, Nicolas~Usunier, and Zagoruyko}]{detr}
Carion, N.; Massa, F.; Synnaeve, G.; Nicolas~Usunier, A.~K.; and Zagoruyko, S. 2020.
\newblock End-to-End Object Detection with Transformers.
\newblock In \emph{ECCV}.

\bibitem[{Changpinyo et~al.(2021)Changpinyo, Sharma, Ding, and Soricut}]{changpinyo2021cc12m}
Changpinyo, S.; Sharma, P.; Ding, N.; and Soricut, R. 2021.
\newblock {Conceptual 12M}: Pushing Web-Scale Image-Text Pre-Training To Recognize Long-Tail Visual Concepts.
\newblock In \emph{CVPR}.

\bibitem[{Chen et~al.(2024{\natexlab{a}})Chen, Li, Dong, Zhang, He, Wang, Zhao, and Lin}]{chen2024sharegpt4v}
Chen, L.; Li, J.; Dong, X.; Zhang, P.; He, C.; Wang, J.; Zhao, F.; and Lin, D. 2024{\natexlab{a}}.
\newblock Sharegpt4v: Improving large multi-modal models with better captions.
\newblock In \emph{European Conference on Computer Vision}, 370--387. Springer.

\bibitem[{Chen et~al.(2024{\natexlab{b}})Chen, Li, Dong, Zhang, Zang, Chen, Duan, Wang, Qiao, Lin et~al.}]{chen2024we}
Chen, L.; Li, J.; Dong, X.; Zhang, P.; Zang, Y.; Chen, Z.; Duan, H.; Wang, J.; Qiao, Y.; Lin, D.; et~al. 2024{\natexlab{b}}.
\newblock Are We on the Right Way for Evaluating Large Vision-Language Models?
\newblock \emph{arXiv:2403.20330}.

\bibitem[{Chiang et~al.(2023)Chiang, Li, Lin, Sheng, Wu, Zhang, Zheng, Zhuang, Zhuang, Gonzalez, Stoica, and Xing}]{vicuna2023}
Chiang, W.-L.; Li, Z.; Lin, Z.; Sheng, Y.; Wu, Z.; Zhang, H.; Zheng, L.; Zhuang, S.; Zhuang, Y.; Gonzalez, J.~E.; Stoica, I.; and Xing, E.~P. 2023.
\newblock Vicuna: An Open-Source Chatbot Impressing GPT-4 with 90\%* ChatGPT Quality.

\bibitem[{Cover and Hart(1967)}]{1053964}
Cover, T.; and Hart, P. 1967.
\newblock Nearest neighbor pattern classification.
\newblock \emph{IEEE Transactions on Information Theory}, 13(1): 21--27.

\bibitem[{Crouse(2023)}]{rectangular_assignment}
Crouse, D. 2023.
\newblock On implementing 2D rectangular assignment algorithms.
\newblock In \emph{IEEE Transactions on Aerospace and Electronic Systems}.

\bibitem[{Duan et~al.(2024)Duan, Yang, Qiao, Fang, Chen, Liu, Dong, Zang, Zhang, Wang et~al.}]{duan2024vlmevalkit}
Duan, H.; Yang, J.; Qiao, Y.; Fang, X.; Chen, L.; Liu, Y.; Dong, X.; Zang, Y.; Zhang, P.; Wang, J.; et~al. 2024.
\newblock Vlmevalkit: An open-source toolkit for evaluating large multi-modality models.
\newblock In \emph{Proceedings of the 32nd ACM International Conference on Multimedia}, 11198--11201.

\bibitem[{Esser, Rombach, and Ommer(2021)}]{esser2021tamingtransformershighresolutionimage}
Esser, P.; Rombach, R.; and Ommer, B. 2021.
\newblock Taming Transformers for High-Resolution Image Synthesis.
\newblock arXiv:2012.09841.

\bibitem[{Fang et~al.(2023)Fang, Wang, Xie, Sun, Wu, Wang, Huang, Wang, and Cao}]{fang2023eva}
Fang, Y.; Wang, W.; Xie, B.; Sun, Q.; Wu, L.; Wang, X.; Huang, T.; Wang, X.; and Cao, Y. 2023.
\newblock Eva: Exploring the limits of masked visual representation learning at scale.
\newblock In \emph{CVPR}.

\bibitem[{Fu et~al.(2024)Fu, Chen, Shen, Qin, Zhang, Lin, Yang, Zheng, Li, Sun, Wu, and Ji}]{fu2024mmecomprehensiveevaluationbenchmark}
Fu, C.; Chen, P.; Shen, Y.; Qin, Y.; Zhang, M.; Lin, X.; Yang, J.; Zheng, X.; Li, K.; Sun, X.; Wu, Y.; and Ji, R. 2024.
\newblock MME: A Comprehensive Evaluation Benchmark for Multimodal Large Language Models.
\newblock arXiv:2306.13394.

\bibitem[{He et~al.(2022)He, Chen, Xie, Li, Doll{\'a}r, and Girshick}]{he2022masked}
He, K.; Chen, X.; Xie, S.; Li, Y.; Doll{\'a}r, P.; and Girshick, R. 2022.
\newblock Masked autoencoders are scalable vision learners.
\newblock In \emph{CVPR}.

\bibitem[{He et~al.(2021)He, Chen, Xie, Li, Dollár, and Girshick}]{he2021maskedautoencodersscalablevision}
He, K.; Chen, X.; Xie, S.; Li, Y.; Dollár, P.; and Girshick, R. 2021.
\newblock Masked Autoencoders Are Scalable Vision Learners.
\newblock arXiv:2111.06377.

\bibitem[{Hondru et~al.(2024)Hondru, Croitoru, Minaee, Ionescu, and Sebe}]{hondru2024maskedimagemodelingsurvey}
Hondru, V.; Croitoru, F.~A.; Minaee, S.; Ionescu, R.~T.; and Sebe, N. 2024.
\newblock Masked Image Modeling: A Survey.

\bibitem[{Hou et~al.(2022)Hou, Sun, Chen, Xie, and Kung}]{hou2022milan}
Hou, Z.; Sun, F.; Chen, Y.-K.; Xie, Y.; and Kung, S.-Y. 2022.
\newblock Milan: Masked image pretraining on language assisted representation.
\newblock \emph{arXiv:2208.06049}.

\bibitem[{Jiang et~al.(2024)Jiang, He, Zeng, Wei, Ku, Liu, and Chen}]{jiang2024mantisinterleavedmultiimageinstruction}
Jiang, D.; He, X.; Zeng, H.; Wei, C.; Ku, M.; Liu, Q.; and Chen, W. 2024.
\newblock MANTIS: Interleaved Multi-Image Instruction Tuning.
\newblock arXiv:2405.01483.

\bibitem[{Jin et~al.(2023)Jin, Xu, Chen, Liao, Tan, Chen, Lei, Liu, Song, Lei et~al.}]{jin2023unified}
Jin, Y.; Xu, K.; Chen, L.; Liao, C.; Tan, J.; Chen, B.; Lei, C.; Liu, A.; Song, C.; Lei, X.; et~al. 2023.
\newblock Unified language-vision pretraining with dynamic discrete visual tokenization.
\newblock \emph{arXiv:2309.04669}.

\bibitem[{Kembhavi et~al.(2016)Kembhavi, Salvato, Kolve, Seo, Hajishirzi, and Farhadi}]{kembhavi2016diagram}
Kembhavi, A.; Salvato, M.; Kolve, E.; Seo, M.; Hajishirzi, H.; and Farhadi, A. 2016.
\newblock A diagram is worth a dozen images.
\newblock In \emph{ECCV}.

\bibitem[{Kuhn(1955)}]{kuhn1955hungarian}
Kuhn, H.~W. 1955.
\newblock The Hungarian method for the assignment problem.
\newblock \emph{Naval research logistics quarterly}.

\bibitem[{Laurençon et~al.(2024)Laurençon, Tronchon, Cord, and Sanh}]{laurençon2024mattersbuildingvisionlanguagemodels}
Laurençon, H.; Tronchon, L.; Cord, M.; and Sanh, V. 2024.
\newblock What matters when building vision-language models?
\newblock arXiv:2405.02246.

\bibitem[{Li et~al.(2023{\natexlab{a}})Li, Wang, Wang, Ge, Ge, and Shan}]{li2023seedbenchbenchmarkingmultimodalllms}
Li, B.; Wang, R.; Wang, G.; Ge, Y.; Ge, Y.; and Shan, Y. 2023{\natexlab{a}}.
\newblock SEED-Bench: Benchmarking Multimodal LLMs with Generative Comprehension.
\newblock arXiv:2307.16125.

\bibitem[{Li et~al.(2024)Li, Zhang, Guo, Zhang, Li, Zhang, Zhang, Zhang, Li, Liu, and Li}]{li2024llavaonevisioneasyvisualtask}
Li, B.; Zhang, Y.; Guo, D.; Zhang, R.; Li, F.; Zhang, H.; Zhang, K.; Zhang, P.; Li, Y.; Liu, Z.; and Li, C. 2024.
\newblock LLaVA-OneVision: Easy Visual Task Transfer.
\newblock arXiv:2408.03326.

\bibitem[{Li et~al.(2023{\natexlab{b}})Li, Li, Savarese, and Hoi}]{li2023blip}
Li, J.; Li, D.; Savarese, S.; and Hoi, S. 2023{\natexlab{b}}.
\newblock Blip-2: Bootstrapping language-image pre-training with frozen image encoders and large language models.
\newblock In \emph{ICML}.

\bibitem[{Li et~al.(2023{\natexlab{c}})Li, Li, Savarese, and Hoi}]{li2023blip2bootstrappinglanguageimagepretraining}
Li, J.; Li, D.; Savarese, S.; and Hoi, S. 2023{\natexlab{c}}.
\newblock BLIP-2: Bootstrapping Language-Image Pre-training with Frozen Image Encoders and Large Language Models.
\newblock arXiv:2301.12597.

\bibitem[{Li et~al.(2023{\natexlab{d}})Li, Wang, Li, Wang, He, Wang, and Qiao}]{li2023unmasked}
Li, K.; Wang, Y.; Li, Y.; Wang, Y.; He, Y.; Wang, L.; and Qiao, Y. 2023{\natexlab{d}}.
\newblock Unmasked teacher: Towards training-efficient video foundation models.
\newblock In \emph{ICCV}.

\bibitem[{Li et~al.(2023{\natexlab{e}})Li, Du, Zhou, Wang, Zhao, and Wen}]{li2023evaluating}
Li, Y.; Du, Y.; Zhou, K.; Wang, J.; Zhao, W.~X.; and Wen, J.-R. 2023{\natexlab{e}}.
\newblock Evaluating object hallucination in large vision-language models.
\newblock \emph{arXiv:2305.10355}.

\bibitem[{Lin et~al.(2024)Lin, Yin, Ping, Molchanov, Shoeybi, and Han}]{lin2024vila}
Lin, J.; Yin, H.; Ping, W.; Molchanov, P.; Shoeybi, M.; and Han, S. 2024.
\newblock Vila: On pre-training for visual language models.
\newblock In \emph{CVPR}.

\bibitem[{Liu et~al.(2023{\natexlab{a}})Liu, Li, Li, and Lee}]{liu2023improvedllava}
Liu, H.; Li, C.; Li, Y.; and Lee, Y.~J. 2023{\natexlab{a}}.
\newblock Improved Baselines with Visual Instruction Tuning.

\bibitem[{Liu et~al.(2024{\natexlab{a}})Liu, Li, Li, and Lee}]{liu2024improvedbaselinesvisualinstruction}
Liu, H.; Li, C.; Li, Y.; and Lee, Y.~J. 2024{\natexlab{a}}.
\newblock Improved Baselines with Visual Instruction Tuning.
\newblock arXiv:2310.03744.

\bibitem[{Liu et~al.(2024{\natexlab{b}})Liu, Li, Li, and Lee}]{liu2024llava_improved}
Liu, H.; Li, C.; Li, Y.; and Lee, Y.~J. 2024{\natexlab{b}}.
\newblock Improved baselines with visual instruction tuning.
\newblock In \emph{CVPR}.

\bibitem[{Liu et~al.(2023{\natexlab{b}})Liu, Li, Wu, and Lee}]{liu2023visualinstructiontuning}
Liu, H.; Li, C.; Wu, Q.; and Lee, Y.~J. 2023{\natexlab{b}}.
\newblock Visual Instruction Tuning.
\newblock arXiv:2304.08485.

\bibitem[{Liu et~al.(2024{\natexlab{c}})Liu, Li, Wu, and Lee}]{liu2024llava}
Liu, H.; Li, C.; Wu, Q.; and Lee, Y.~J. 2024{\natexlab{c}}.
\newblock Visual instruction tuning.
\newblock In \emph{NeurIPS}.

\bibitem[{Liu et~al.(2024{\natexlab{d}})Liu, Duan, Zhang, Li, Zhang, Zhao, Yuan, Wang, He, Liu et~al.}]{liu2023mmbench}
Liu, Y.; Duan, H.; Zhang, Y.; Li, B.; Zhang, S.; Zhao, W.; Yuan, Y.; Wang, J.; He, C.; Liu, Z.; et~al. 2024{\natexlab{d}}.
\newblock Mmbench: Is your multi-modal model an all-around player?
\newblock \emph{ECCV}.

\bibitem[{Liu et~al.(2024{\natexlab{e}})Liu, Li, Huang, Yang, Yu, Li, Yin, Liu, Jin, and Bai}]{Liu_2024}
Liu, Y.; Li, Z.; Huang, M.; Yang, B.; Yu, W.; Li, C.; Yin, X.-C.; Liu, C.-L.; Jin, L.; and Bai, X. 2024{\natexlab{e}}.
\newblock OCRBench: on the hidden mystery of OCR in large multimodal models.
\newblock \emph{Science China Information Sciences}, 67(12).

\bibitem[{Loshchilov and Hutter(2019)}]{loshchilov2017decoupled}
Loshchilov, I.; and Hutter, F. 2019.
\newblock Decoupled Weight Decay Regularization.
\newblock In \emph{ICLR}.

\bibitem[{Mizrahi et~al.(2023)Mizrahi, Bachmann, Kar, Yeo, Gao, Dehghan, and Zamir}]{mizrahi20234mmassivelymultimodalmasked}
Mizrahi, D.; Bachmann, R.; Kar, O.~F.; Yeo, T.; Gao, M.; Dehghan, A.; and Zamir, A. 2023.
\newblock 4M: Massively Multimodal Masked Modeling.
\newblock In \emph{NeurIPS}.

\bibitem[{Papadimitriou and Steiglitz(1998)}]{hungarian_papad}
Papadimitriou, C.~H.; and Steiglitz, K. 1998.
\newblock \emph{Combinatorial optimization: algorithms and complexity}.
\newblock Courier Corporation.

\bibitem[{Qwen et~al.(2025)Qwen, :, Yang, Yang, Zhang, Hui, Zheng, Yu, Li, Liu, Huang, Wei, Lin, Yang, Tu, Zhang, Yang, Yang, Zhou, Lin, Dang, Lu, Bao, Yang, Yu, Li, Xue, Zhang, Zhu, Men, Lin, Li, Tang, Xia, Ren, Ren, Fan, Su, Zhang, Wan, Liu, Cui, Zhang, and Qiu}]{qwen2025qwen25technicalreport}
Qwen; :; Yang, A.; Yang, B.; Zhang, B.; Hui, B.; Zheng, B.; Yu, B.; Li, C.; Liu, D.; Huang, F.; Wei, H.; Lin, H.; Yang, J.; Tu, J.; Zhang, J.; Yang, J.; Yang, J.; Zhou, J.; Lin, J.; Dang, K.; Lu, K.; Bao, K.; Yang, K.; Yu, L.; Li, M.; Xue, M.; Zhang, P.; Zhu, Q.; Men, R.; Lin, R.; Li, T.; Tang, T.; Xia, T.; Ren, X.; Ren, X.; Fan, Y.; Su, Y.; Zhang, Y.; Wan, Y.; Liu, Y.; Cui, Z.; Zhang, Z.; and Qiu, Z. 2025.
\newblock Qwen2.5 Technical Report.
\newblock arXiv:2412.15115.

\bibitem[{Radford et~al.(2021)Radford, Kim, Hallacy, Ramesh, Goh, Agarwal, Sastry, Askell, Mishkin, Clark et~al.}]{radford2021learning}
Radford, A.; Kim, J.~W.; Hallacy, C.; Ramesh, A.; Goh, G.; Agarwal, S.; Sastry, G.; Askell, A.; Mishkin, P.; Clark, J.; et~al. 2021.
\newblock Learning transferable visual models from natural language supervision.
\newblock In \emph{ICML}.

\bibitem[{Tong et~al.(2024)Tong, Brown, Wu, Woo, Middepogu, Akula, Yang, Yang, Iyer, Pan, Wang, Fergus, LeCun, and Xie}]{tong2024cambrian1fullyopenvisioncentric}
Tong, S.; Brown, E.; Wu, P.; Woo, S.; Middepogu, M.; Akula, S.~C.; Yang, J.; Yang, S.; Iyer, A.; Pan, X.; Wang, Z.; Fergus, R.; LeCun, Y.; and Xie, S. 2024.
\newblock Cambrian-1: A Fully Open, Vision-Centric Exploration of Multimodal LLMs.
\newblock arXiv:2406.16860.

\bibitem[{Touvron et~al.(2023)Touvron, Lavril, Izacard, Martinet, Lachaux, Lacroix, Rozi{\`e}re, Goyal, Hambro, Azhar et~al.}]{touvron2023llama}
Touvron, H.; Lavril, T.; Izacard, G.; Martinet, X.; Lachaux, M.-A.; Lacroix, T.; Rozi{\`e}re, B.; Goyal, N.; Hambro, E.; Azhar, F.; et~al. 2023.
\newblock Llama: Open and efficient foundation language models.
\newblock \emph{arXiv:2302.13971}.

\bibitem[{Tschannen et~al.(2025)Tschannen, Gritsenko, Wang, Naeem, Alabdulmohsin, Parthasarathy, Evans, Beyer, Xia, Mustafa, Hénaff, Harmsen, Steiner, and Zhai}]{tschannen2025siglip2multilingualvisionlanguage}
Tschannen, M.; Gritsenko, A.; Wang, X.; Naeem, M.~F.; Alabdulmohsin, I.; Parthasarathy, N.; Evans, T.; Beyer, L.; Xia, Y.; Mustafa, B.; Hénaff, O.; Harmsen, J.; Steiner, A.; and Zhai, X. 2025.
\newblock SigLIP 2: Multilingual Vision-Language Encoders with Improved Semantic Understanding, Localization, and Dense Features.
\newblock arXiv:2502.14786.

\bibitem[{Wang et~al.(2024)Wang, Zheng, Zhao, Wang, Ge, Zhang, and Zhang}]{wang2024reconstructivevisualinstructiontuning}
Wang, H.; Zheng, A.; Zhao, Y.; Wang, T.; Ge, Z.; Zhang, X.; and Zhang, Z. 2024.
\newblock Reconstructive Visual Instruction Tuning.
\newblock arXiv:2410.09575.

\bibitem[{Wang et~al.(2023)Wang, Lv, Yu, Hong, Qi, Wang, Ji, Yang, Zhao, Song et~al.}]{wang2023cogvlm}
Wang, W.; Lv, Q.; Yu, W.; Hong, W.; Qi, J.; Wang, Y.; Ji, J.; Yang, Z.; Zhao, L.; Song, X.; et~al. 2023.
\newblock Cogvlm: Visual expert for pretrained language models.
\newblock \emph{arXiv:2311.03079}.

\bibitem[{Wei et~al.(2022)Wei, Xie, Zhou, Li, and Tian}]{wei2022mvp}
Wei, L.; Xie, L.; Zhou, W.; Li, H.; and Tian, Q. 2022.
\newblock Mvp: Multimodality-guided visual pre-training.
\newblock In \emph{ECCV}.

\bibitem[{xAI(2024)}]{grok1.5v}
xAI. 2024.
\newblock Grok-1.5V.
\newblock \url{https://x.ai/news/grok-1.5v}.
\newblock Accessed: 2025-07-23.

\bibitem[{Xie et~al.(2024)Xie, Mao, Bai, Zhang, Wang, Lin, Gu, Chen, Yang, and Shou}]{xie2024showosingletransformerunify}
Xie, J.; Mao, W.; Bai, Z.; Zhang, D.~J.; Wang, W.; Lin, K.~Q.; Gu, Y.; Chen, Z.; Yang, Z.; and Shou, M.~Z. 2024.
\newblock Show-o: One Single Transformer to Unify Multimodal Understanding and Generation.
\newblock arXiv:2408.12528.

\bibitem[{Yang et~al.(2023)Yang, Ge, Yi, Li, Shan, Qie, and Wang}]{yang2023rils}
Yang, S.; Ge, Y.; Yi, K.; Li, D.; Shan, Y.; Qie, X.; and Wang, X. 2023.
\newblock Rils: Masked visual reconstruction in language semantic space.
\newblock In \emph{CVPR}.

\bibitem[{Yue et~al.(2024)Yue, Ni, Zhang, Zheng, Liu, Zhang, Stevens, Jiang, Ren, Sun, Wei, Yu, Yuan, Sun, Yin, Zheng, Yang, Liu, Huang, Sun, Su, and Chen}]{yue2024mmmumassivemultidisciplinemultimodal}
Yue, X.; Ni, Y.; Zhang, K.; Zheng, T.; Liu, R.; Zhang, G.; Stevens, S.; Jiang, D.; Ren, W.; Sun, Y.; Wei, C.; Yu, B.; Yuan, R.; Sun, R.; Yin, M.; Zheng, B.; Yang, Z.; Liu, Y.; Huang, W.; Sun, H.; Su, Y.; and Chen, W. 2024.
\newblock MMMU: A Massive Multi-discipline Multimodal Understanding and Reasoning Benchmark for Expert AGI.
\newblock arXiv:2311.16502.

\bibitem[{Zhu et~al.(2023)Zhu, Chen, Shen, Li, and Elhoseiny}]{zhu2023minigpt}
Zhu, D.; Chen, J.; Shen, X.; Li, X.; and Elhoseiny, M. 2023.
\newblock Minigpt-4: Enhancing vision-language understanding with advanced large language models.
\newblock \emph{arXiv:2304.10592}.

\end{thebibliography}

\end{document}